\documentclass[
]{ceurart}

\usepackage{listings}
\usepackage{amsthm}
\theoremstyle{definition}
\newtheorem{theorem}{Definition}
\usepackage[acronym, symbols]{glossaries}
\makeglossaries
\newacronym{auc}{AUC}{Area under the receiver operating characteristic curve}
\newacronym{dgg}{DGG}{Discrete Global Grid}
\newacronym{kg}{KG}{Knowledge Graph}
\newacronym{osm}{OSM}{OpenStreetMap}
\newacronym{stkg}{STKG}{Spatial-Temporal Knowledge Graph}
\newacronym{id}{ID}{Identifier}
\newacronym{de9im}{DE-9IM}{Dimensionally Extended 9-Intersection Model}
\newacronym{pca}{PCA}{Principal Component Analysis}
\newacronym{mae}{MAE}{Mean Absolute Error}
\newacronym{mape}{MAPE}{Mean Absolute Percentage Error}
\begin{document}
\copyrightyear{2024}
\copyrightclause{Copyright for this paper by its authors.
  Use permitted under Creative Commons License Attribution 4.0
  International (CC BY 4.0).}

\conference{GeoLD2024: 6th Geospatial Linked Data Workshop,
  May 26, 2024, Hersonissos, Greece}

\title{A Planet Scale Spatial-Temporal Knowledge Graph Based On OpenStreetMap And H3 Grid}

\author[1]{Martin B\"ockling}[%
orcid=0000-0002-1143-4686,
email=martin.boeckling.gast@uni-mannheim.de
]
\cormark[1]
\address[1]{Data and Web Science Group University of Mannheim,
  B6 26, Mannheim, 68159, Germany}
\address[2]{Corporate State University of Mannheim,
  Colblitzallee 1-9, Mannheim, 68163, Germany}

\author[1]{Heiko Paulheim}[%
orcid=0000-0003-4386-8195,
email=heiko@informatik.uni-mannheim.de
]

\author[2]{Sarah Detzler}[%
orcid=0000-0002-7504-8856,
email=sarah.detzler@dhbw-mannheim.de
]

\cortext[1]{Corresponding author.}

\begin{abstract}
  Geospatial data plays a central role in modeling our world, for which OpenStreetMap (OSM) provides a rich source of such data. While often spatial data is represented in a tabular format, a graph based representation provides the possibility to interconnect entities which would have been separated in a tabular representation. We propose in our paper a framework which supports a planet scale transformation of OpenStreetMap data into a Spatial Temporal Knowledge Graph. In addition to OpenStreetMap data, we align the different OpenStreetMap geometries on individual h3 grid cells. We compare our constructed spatial knowledge graph to other spatial knowledge graphs and outline our contribution in this paper. As a basis for our computation, we use Apache Sedona as a computational framework for our Spatial Temporal Knowledge Graph construction.
\end{abstract}


\maketitle

\section{Introduction}
Within the spatial domain \gls*{osm} provides a large extent of open-source spatial data that is modeled by different contributors. It provides data annotated by different contributors and official data providers, which represents data over the entire planet \cite{openstreetmap_contributors_openstreetmap_2024}. Often spatial data is represented in a tabular format. Graphs, but in particular \glspl*{kg}, provide a good foundation to interconnect related entities in the spatial domain. They can model entities and events in a multi-faceted way, where distance in the graph can express both geographic as well as semantic distances. When embedding such a graph using knowledge graph embedding methods, it is possible to create latent spaces where both facets, geographic and semantic proximity, are jointly reflected.

For our research paper, we focus on the spatial data source provided by \gls*{osm}. We use the h3 \gls*{dgg} to provide a regularization for the different \gls*{osm} geometries, so that each individual cell tessellates the earth uniquely. This does not only involve data from a one-time snapshot, but we aim at providing a \gls*{kg} that is modeled over a temporal dimension. In the following sections, we outline our approach for a scalable \gls*{stkg} over the entire planet. Furthermore, we conclude our research by comparing it conceptually to other \glspl*{stkg}.

\section{Related Work} \label{cha:RelatedWork}
Within the spatial data domain, various data representations for graph-based datasets are feasible. One representation involves static nodes and edge pairs. Static graphs, exemplified by road networks or electricity grids \cite{barthelemy_spatial_2011}, are a specific instance of this data representation. The capability to establish connections between spatial information through a graph enables the modeling of links between individual geometries or multiple geometries, providing a means to represent complex systems \cite{morris_transport_2012}.

In the realm of \glspl*{kg} within the spatial domain, our focus is directed towards a curated selection of diverse \glspl*{kg}. In this section we will highlight a selection of \glspl*{kg}, but we will not provide an extensive list of different \glspl*{kg}. One notable \gls*{kg}, WorldKG, is specifically designed to encapsulate \gls*{osm} data. The structural organization of WorldKG involves the transposition of features extracted from \gls*{osm}, establishing relationships between various categories. In this \gls*{kg}, tags derived from \gls*{osm} function as separate entities, framing the hierarchical structure of WorldKG. Each individual feature extracted from \gls*{osm} is represented as a distinct node within the WorldKG. Moreover, the geometries associated with each \gls*{osm} entity are also exposed as point geometry as nodes within the WorldKG. However, no grid is used to represent geographical areas. In total, the published \gls*{kg} is based on \gls*{osm} data from June 06, 2021. In total, the published \gls*{kg} contains over 828 million triples and over 113 million entities, categorized into 33 different top-level classes. \cite{dsouza_worldkg_2021}

In comparison to WorldKG, the KnowWhereGraph framework is constructed upon a diverse array of datasets encompassing hazard information, climate data, soil properties, crop and land cover types, as well as demographic and human health data. To facilitate the integration of these heterogeneous datasets, the framework employs the S2 discrete hierarchical grid, thereby harmonizing location data from various sources. The S2 grid uses squared shapes as grid cells, and the hierarchy allows larger and smaller granularity. Each discrete hierarchical grid cell functions as a unique identifier for the corresponding region. In conjunction to the grid cell \gls*{id}, various other regional attributes, such as ZIP codes, administrative regions, or Climate Division Boundaries, are systematically mapped to the respective areas. During the time of the publication, the Knowledge Graph consisted of 4.9 billion triples. \cite{janowicz_know_2022}

The current approaches for constructing spatial Knowledge Graphs come with certain shortcomings. For instance, WorldKG currently considers only Point geometry types as an input. While providing a semantically enriched representation for \gls*{osm} metadata, geographic near elements are not necessarily close in the graph. In comparison, KnowWhereGraph uses other data sources limited to selected datasets from the United States. It provides in addition relations between geometries and the \gls*{dgg}. 

In this paper, we propose a framework that allows us to transform \gls*{osm} data into a \gls*{kg}. Core considerations for our approach are the involvement of a \gls*{dgg} together with a modeling of the relations between the geometry and an individual grid cell. Furthermore, the \gls*{kg} should provide a temporal dimension to the dataset. In the following sections, we provide an overview of the theoretical background of spatial data, as well as the transformation into a \gls*{kg}.

\section{Theoretical Background of Spatial Data and Knowledge Graphs}
For the following sections we introduce the related concepts for our \gls*{kg} with concepts from the spatial domain. Furthermore, we will provide an introduction to the data structure  of \gls*{osm}. 
\subsection{OpenStreetMap}
\gls*{osm} provides a data foundation that allows to map geographical entities. The complete project is open source and therefore provides a unique opportunity in the spatial domain for data analysis. \gls*{osm} provides two different types of data categories. The first data category involves the so-called \gls*{osm} map features. Map features provide additional information to \gls*{osm} entities and represent geographical attributes. Map features are modeled by using key-value pairs, in which the key models the primary feature and the value further specifies the primary feature. We define \gls*{osm} map features as followed \cite{openstreetmap_contributors_map_2023}:
\begin{theorem}
    Let $K$ be the set of all possible keys and $V$ the set of all possible values. We therefore denote all tags as $T$. Therefore, \gls*{osm} map features are represented as $T: K \to V$. 
\end{theorem}

For instance, the key \textit{highway} can be specified with the value \textit{motorway}. The map feature \textit{highway=motorway} describes therefore a divided highway with more than two lanes. \gls*{osm} does not restrict the variety of map features that can be defined, which also provides the possibility to have multilingual map features defined by users. Nevertheless, \gls*{osm} provides commonly accepted map features \cite{openstreetmap_contributors_map_2023}.

The second data category is described as \gls*{osm} elements. They are divided into Nodes, Ways and Relations. The Node element represents a point within a geographic space. Nodes represent its geographic location as a pair of longitude and latitude. Nodes can represent, for instance, coffee shops, trees, or park benches. We use for the Node element the following definition:

\begin{theorem}
    Let $A_{node}$ be the set of \gls*{osm} Node attribute names and let $W_{node}$ be the possible values for the attributes $A_{node}$. We define properties $P_{node}$ as $P_{node}: A_{node} \to W_{node}$. For node objects $N_{node}$ we define the following attributes $A_{node}=\{"id", "version", "changeset", "lat", "lon", "user", "uid", "visible", "timestamp"\}$. A node $O_{node}$ consists of the following tuple $O_{node}=(P_{node} , T)$.
\end{theorem}

Ways represent in the context of \gls*{osm} linear geometries. In general, Ways can be translated into Line or Polygon geometries. A Way within \gls*{osm} consists of a set of Node elements. The combination of the Point location from nodes constructs the Line or Polygon geometries. Objects that use Line geometries are highways or power lines. Polygons encode, for example, buildings or land-use areas. The data structure of a \gls*{osm} Way is defined as followed:
\begin{theorem}
    Let $A_{way}$ be the set of \gls*{osm} way attribute names and let $W_{way}$ be the possible values for the attributes $A_{way}$. We define $P_{way}$ as $P_{way}: A_{way} \to W_{way}$. We define $N$ to be the set of of node references in a way. Each element in $N$ references a node identified by an ID. For way objects $O_{way}$ we define the following attributes $A_{way}=\{"id", "version", "changeset", "user", "uid", "timestamp"\}$. A way object $O_{way}$ consists of the following tuple $O_{way}=(P_{way}, N, T)$.
\end{theorem}
The third element of the \gls*{osm} data structure is a relation. Relations consist of Nodes, Ways, or other Relations to define logical geographical relations to the different elements. Each element within a relation specifies a role element, which represents the function of the element within the relation. Those generally represent administrative boundaries and routes. The definition of an \gls*{osm} relation is defined as followed:

\begin{theorem}
    Let $A_{relation}$ be the set of \gls*{osm} relation attribute names and let $W_{relation}$ be the possible values for the attributes $A_{relation}$. We define the relation property $P_{relation}$ as $P_{relation}: A_{relation} \to W_{relation}$. Within a relation we define members $M$ that either represent Nodes or Ways. Each member is consisting of a tuple of type, reference and role. For $O_{relation}$ we define following attributes $A_{relation} = \{"id", "version", "changeset", "user", "uid", "timestamp"\}$. We define a relation object $O_{relation}$ as a tuple consisting of relation properties $P_{relation}$, members $M$ and tags $T$: $R_{relation} = (P_{relation}, M, T)$.
\end{theorem}

Based on the data structure of \gls*{osm}, a geometry for an \gls*{osm} entity cannot be retrieved directly. Instead, the geometry for each of the three different data structures needs to be constructed explicitly. In the following subsection \ref{cha:SpatialFoundation} the theoretical foundation for the used spatial methods are outlined.

\subsection{Spatial Foundation}\label{cha:SpatialFoundation}
The geometries that are extracted from \gls*{osm} can range from simple point geometries to more complex geometry representations like Polygons or Geometry Collections. To build up the relationships between the different geographic relationships, the topological model \gls*{de9im} is used. Similarly to our methodology, KnowWhereGraph also uses the \gls*{de9im} methodology to model the geometry-grid relationship \cite{janowicz_know_2022}. Furthermore, different scientific papers use the DE-9IM method to model the spatial relationship between spatial geometries \cite{romanschek_novel_2021}. 

The \gls*{de9im} models the relationships between geometries by using a 3x3-dimensional matrix. It models the relationship between the different geometries by intersecting the interior $I$, exterior $E$ and boundary $B$ of two geometries. Assuming two geometries $a$ and $b$, the dimension of the intersection between geometric objects $a$ and $b$ can be calculated with the function \text{dim}. The dimension function dim of a general set of geometry $S$ returns for relation determination the highest value, and is defined as followed \cite{clementini_modelling_1994}:

\begin{equation} \label{eq:DimFunction}
    \text{dim}(S)=
    \begin{cases}
		-1 & \text{if $S$ is empty}\\
		0 & \text{if $S$ contains a point and no lines or areas}\\
		1 & \text{if $S$ contains a line and no areas}\\
		2 & \text{if $S$ contains an area}
	\end{cases}.
\end{equation}

The complete intersection matrix on which the \gls*{de9im} builds up is given in \autoref{eq:DE9IM}. The \gls*{de9im} method intersects the two geometric objects a and b for each of the topological properties of $I$, $B$ and $E$. Based on the intersection of both geometries, the dimension function $dim$ is applied. \cite{clementini_modelling_1994}

\begin{equation} \label{eq:DE9IM}
	\text{DE9IM}(a,b) = 
	\begin{bmatrix}
		dim(I(a)\cap I(b)) & dim(I(a)\cap B(b))  & dim(I(a)\cap E(b)) \\
		dim(B(a)\cap I(b)) & dim(B(a)\cap B(b)) & dim(B(a)\cap E(b)) \\
		dim(E(a)\cap I(b)) & dim(E(a)\cap B(b)) & dim(E(a)\cap E(b)) \\
	\end{bmatrix}
\end{equation}

When applying the \gls*{de9im} model to two geometries, the 3×3 matrix is filled with the respective result of the $dim$ function. By concatenating the output of the \gls*{de9im} associated spatial predicates can be determined. For the determination of the spatial predicates, the value range from the function dim displayed in \autoref{eq:DimFunction} can be masked with the symbol set {T, F *}. Element T masks dim values {0, 1, 2}, element F masks the return value {-1}, and element * masks the values {-1, 0, 1, 2}. When applying the \gls*{de9im} method, the results can be merged into a string by concatenating element-wise the results from the method. In \autoref{tab:de9im} a selection of the \gls*{de9im} patterns are shown to predicates \cite{feng_discovery_2010}.

\begin{table}[ht]
    \centering
    \caption{Spatial predicate and associated \gls*{de9im} pattern \cite{feng_discovery_2010}}
    \label{tab:de9im}
    \begin{tabular}{p{0.3\textwidth}p{0.3\textwidth}p{0.3\textwidth}}
    	\toprule
    	Spatial predicate & Spatial geometry combination & DE-9IM pattern\\
    	\midrule
    	Equals & All & T*F**FFF*\\
    	Disjoint & All & FF*FF****\\
    	Touches & All except Point $\cap$ Point & FT******* $\vee$ F**T***** $\vee$ F***T****\\
    	Crosses & Point $\cap$ Line & T*T******\\
    	Crosses & Line $\cap$ Line & 0********\\
    	Within & All & T*F**F***\\
    	Overlaps & Point $\cap$ Point, Area $\cap$ Area & T*T***T**\\
    	Overlaps & Line $\cap$ Line & 1*T***T**\\
    	Intersects & All & a.intersects(b) =$\neg$b.disjoint(a)\\
    	Contains & All & a.contains(b) = b.within(a) \\
    	\bottomrule
    \end{tabular}
\end{table}

As a basis for our \gls*{stkg}, we use a \gls*{dgg} to harmonize our geometries that are extracted from \gls*{osm} similar to KnowWhereGraph \cite{janowicz_know_2022}. With \glspl*{dgg} we have the possibility to extract global unique \glspl*{id} per individual grid cell which allows for extensibility of the \gls*{kg}. Instead of using the S2 \gls*{dgg}, we will base our geometry relation on the h3 \gls*{dgg}. The most important difference is that the S2 \gls*{dgg} uses a square-based grid cell geometry, whereas h3 uses a hexagonal-based grid cell geometry \cite{li_geospatial_2020}. The main reason for our decision is that in a hexagonal grid, the distance to all neighboring cells is uniform, which is not the case for square- or triangular-based grids.

\section{Outline of Knowledge Graph representation} \label{cha:Ontology}
For our \gls*{kg} we provide a representation where the geometries of a geographic entity are aligned on a \gls*{dgg} and modeled over time. Besides the geographic relation also the tag information provided by \gls*{osm} is encoded within the \gls*{kg}. Overall, the representation for the \gls*{kg} ontology should respect the outlined rules:
\begin{itemize}
    \item Represent spatial entities from \gls*{osm} in quadruple structure
    \item Include hierarchical relations of common \gls*{osm} tags in \gls*{kg}
    \item Provide spatial relationship between \gls*{osm} geometries and \gls*{dgg}
    \item Provide relationships between individual grid cells in \gls*{dgg}
\end{itemize}

\subsection{Overview of Classes and Properties}
Starting with the commonly used \gls*{osm} tags, we expose the \gls*{osm} entity in our ontology  using the \gls*{osm} \gls*{id}. Between the common tags, we build a subclass relation based on the respective hierarchy in the \gls*{osm} tags. For the spatial relationships, we adapt the GeoSPARQL classes and properties to align the created \gls*{kg} to the respective standard. The individual object of \gls*{osm} has for the respective value the relation \textit{rdf:type}. If an \gls*{osm} tag is not within the commonly addressed tags, we use the key value pair within our \gls*{kg}. The \gls*{osm} \gls*{id} is the subject, the key acts as the predicate and the value is used as the object. 

\begin{figure}[ht]
    \centering
    \includegraphics[width=0.4\textwidth]{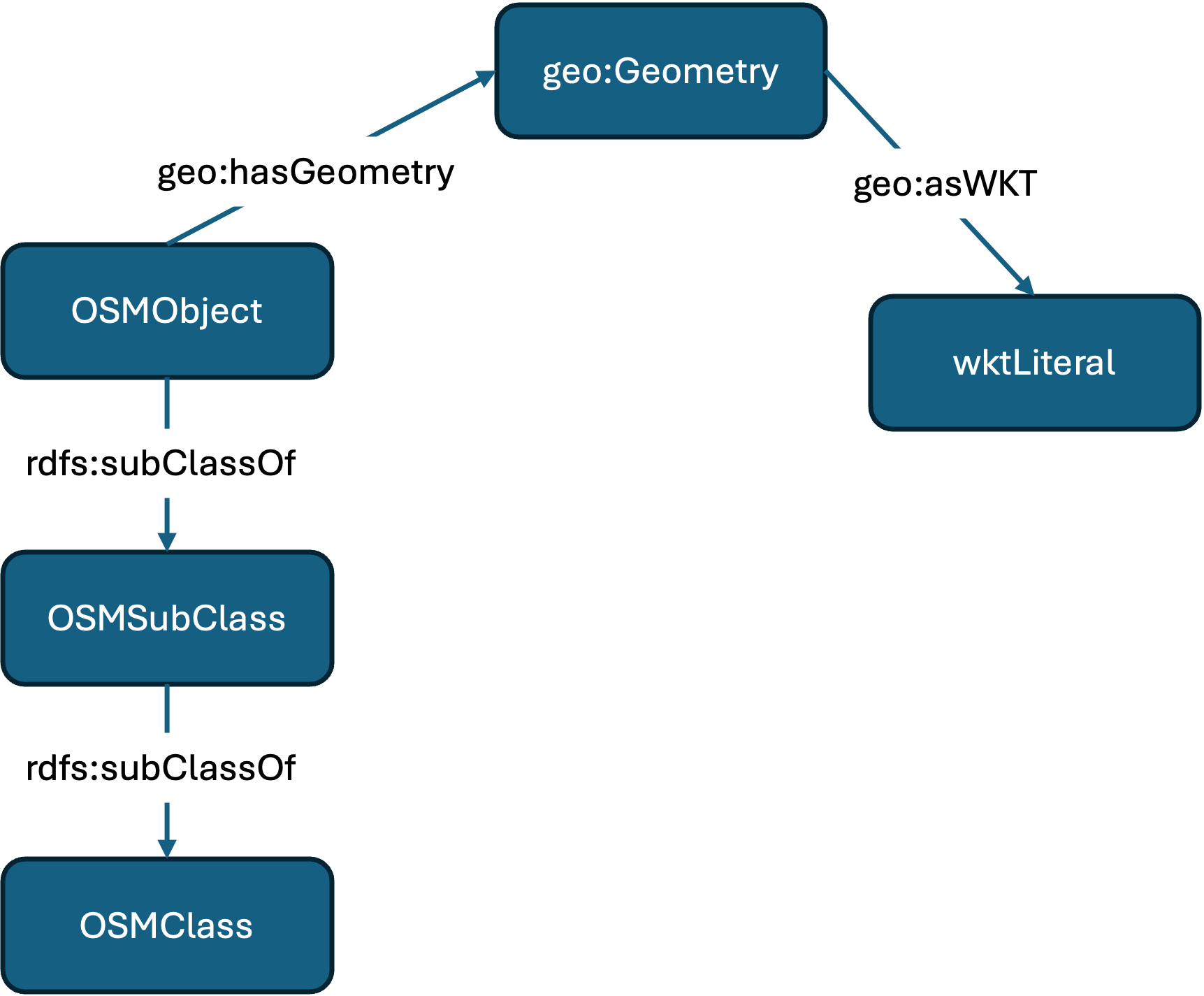}
    \caption{Ontology of our Spatial-Temporal Knowledge Graph for OpenStreetMap objects}
    \label{fig:OSMOntology}
\end{figure}

For the different individual grid cells, we use the related \gls*{de9im} GeoSPARQL properties to model the relation to \gls*{osm}. This is based on the conditions outlined in table \ref*{tab:de9im}. This involves the following properties: \textit{geo:sfContains}, \textit{geo:sfCrosses}, \textit{geo:sfEquals}, \textit{geo:sfOverlaps}, \textit{geo:sfTouches}, \textit{geo:sfWithin}, \textit{geo:ehCovers}, \textit{geo:ehCoveredBy}, and \textit{geo:sfIntersects}. Due to the hierarchical relationships between the \gls*{de9im} methodology, multiple relationships can be directed from the grid cell to the individual \gls*{osm} geometry. An example of how for one specific \gls*{osm} \gls*{id} the \gls*{stkg} is structured is outlined in subsection \ref{cha:KGExample}.


Within our \gls*{stkg}, we use a \gls*{dgg}, on which we align the \gls*{osm} geometries for the \gls*{kg}. Specifically for \gls*{dgg}, an ontology has already been proposed. Similar to our ontology, for the relation to other geometries not from the grid, the GeoSPARQL classes have been adopted using, for instance, the \gls*{de9im} methodology. However, for the selected h3 grid, the ontology can only be partially used, as the only relations for the grid cells are \textit{hcf:isAdjacentTo} and \textit{hcf:contains}, which are based on the \gls*{de9im} contains predicate \cite{shimizu_pattern_2021}. While this works for square-based grid systems, as grid cells on different hierarchy levels always contain the respective smaller grid cell, hexagonal grid cells or triangular grid cells on different resolutions do not fulfill this argument. We therefore expand the hierarchical relationships properties to \textit{isParentCellOf} and \textit{isChildCellOf} to capture the hierarchical relationship between grid cells.

\begin{figure}[ht]
    \centering
    \includegraphics[width=0.9\textwidth]{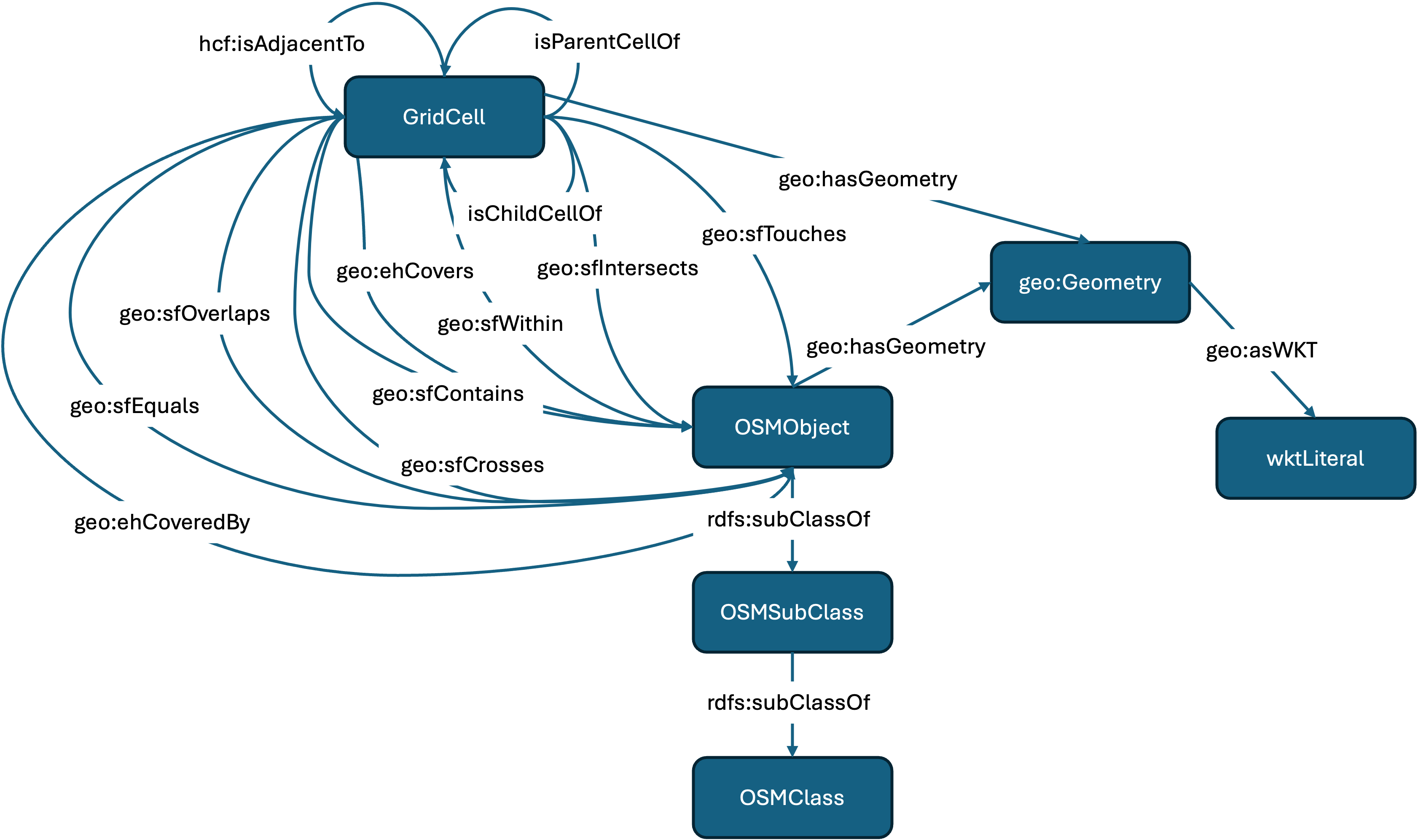}
    \caption{Ontology of our Spatial-Temporal Knowledge Graph for grid cell ontology}
    \label{fig:OSMGrid}
\end{figure}

\subsection{Example of Knowledge Graph Entity} \label{cha:KGExample}
To display how the structure of the \gls*{stkg} looks, we use a sample \gls*{osm} object and display the respective part of the \gls*{stkg}. In addition to the \gls*{osm} object, we also display the related h3 grid cell to the \gls*{osm} object. As an example entity for \gls*{osm}, we use \gls*{osm} Way 240974013. In total, we transform the single element of the \gls*{osm} \gls*{id} and the related h3 grid cell with its neighbors into 51 quadruples. As the complete list of quadruples would be too long for the paper, we publish a full example under the following \href{https://github.com/MartinBoeckling/stkg_construction/blob/main/kg_example/kg_data/part-00000-2a93b988-d93f-4f13-b796-19f5784e4bf4-c000.csv}{link of our Github}. In table \ref{tab:STKGExample}, we provide an excerpt from the complete example.

\begin{table}[ht]
    \centering
    \caption{Example of our Spatio-Temporal Knowledge Graph for the University building of Mannheim, represented in Way 240974013}
    \label{tab:STKGExample}
    \begin{tabular}{p{0.2\textwidth}p{0.2\textwidth}p{0.2\textwidth}p{0.2\textwidth}}
    	\toprule
    	subject & predicate & object & date\\
    	\midrule
        240974013 & rdf:type & university & 2024-01-01 \\
        university & rdfs:subClassOf & amenity & 2024-01-01 \\
        240974013 & addr:city & Mannheim & 2024-01-01 \\
        240974013 & geo:hasGeometry & geo240974013 & 2024-01-01 \\
        geo240974013 & geo:asWKT & POLYGON (([...])) & 2024-01-01 \\
        881fae61b9fffff & geo:sfContains & 240974013 & 2024-01-01 \\
        881fae61b9fffff & geo:ehCovers & 240974013 & 2024-01-01 \\
        881fae61b9fffff & geo:sfIntersects & 240974013 & 2024-01-01 \\
    	\bottomrule
    \end{tabular}
\end{table}

\section{Knowledge Graph Data Preparation}
For the data preparation to construct our \gls*{stkg}, we used the \gls*{osm} files from Geofabrik. To provide an overview of the implemented approach, we have divided our data preparation into three different phases: (1) the specific \gls*{osm} data preparation, (2) the h3 \gls*{dgg} data preparation, and (3) the \gls*{stkg} construction. Throughout the data preparation phase, we use Parquet files as the main data storage format. It provides the best trade-off between data storage costs and possibilities to push down queries to the file itself \cite{hu_compared_2018}. The capability of predicate push downs to Parquet files allows during the data preparation phase to only scan the relevant data for the necessary transformation \cite{luo_batch_2022, hu_compared_2018}.

\subsection{Preparation of OpenStreetMap data}
We use \gls*{osm} as a data foundation for our \gls*{stkg}. Specifically, the \textit{.osm.pbf} files are used, which can be derived from Geofabrik\footnote{Geofabrik provides \gls*{osm} data extracts in different formats under the following \href{http://download.geofabrik.de}{web page}}. Each \gls*{osm} file represents a specified time stamp for a specific region. For the three different main \gls*{osm} data structures, we need to convert the XML-based structure into a table-based structure for the future processing of the data. For the Node data structure, the coordinates are given, for all other data structures, the geometries need to be explicitly constructed. In order to achieve that, we use a gdal-based method called ogr2ogr, which allows us to natively read .osm.pbf files and convert them to other formats while constructing the respective geometries for the different formats. In our case, we use ogr2ogr to convert the .osm.pbf files to Parquet files to further process them for our \gls*{stkg}. For that, each.osm.pbf is split into five different files, which address the different layers gdal assigns to \gls*{osm} files.

After the initial conversion of the .osm.pbf file, we use the five resulting files to optimize them further for the future \gls*{stkg} construction. For that, we use Apache Sedona as the main processing method for the \gls*{stkg}. In comparison to other spatial data frameworks, Apache Sedona has shown a better speed and memory efficiency when using it in computational extensive workloads \cite{tahboub_architecting_2020,pandey_how_2018}. Therefore, we decided to use Apache Sedona as our processing engine. With Apache Sedona, the possibility exists to write Parquet files that support predicate pushdown for geometric operations. To support the spatial predicate pushdown, the files need to contain, for each geometry, a geohash and then be written back to a Parquet file. From the file name we further extract the specific date the \gls*{osm} file represents and add the date as a column to the written file. Based on the transformed geometry files, we build up our grid, where we will use the h3 grid system. In subsection \ref{cha:h3Grid}, we outline the approach to grid creation for our \gls*{stkg}.
\subsection{h3 Grid Data Preparation} \label{cha:h3Grid}
Based on the requirements outlined in section \ref{cha:Ontology}, we generate the h3 grid, which is used to align geometries on it. For the respective grid cells, we use a world map as an input to retrieve the h3 grid cells using the h3 Python package \cite{uber_technologies_h3-py_2023}. Depending on the parameter, we use the Nominatim API to create a bounding box and clip the individual geometries of the world map. This allows to not only construct a world-wide h3 grid, but also region-specific \gls*{dgg}.

For the h3 grid we allow to specify grid-specific parameters. This includes the definition of the resolution for each individual grid cell. The values that could be used are in the range of $[0, 15] \subseteq \mathbb{N}$, where 0 represents the resolution of the biggest cell area with an average size of 4,357,449.41 km$^2$. For the resolution of 15, an individual h3 grid cell has an average area size of 0.895 m$^2$, which is the smallest grid cell area captured in the h3 grid. Additionally, the h3 grid system allows the compacting of all grid cells based on population density. This allows, especially for sparsely populated regions, to represent the data in fewer grid cells while still representing the same area. After receiving all the respective grid cell \glspl*{id}, we retrieve for each individual grid cell the respective geometry associated with the individual grid cell. After retrieving for all specified regions all grid cells, we store the grid data in a Parquet file. Similar to \gls*{osm}, we use Apache Sedona to store the geohash per individual grid cell to optimize the predicate push down to the Parquet file. After finishing the grid creation, we outline in the last step of our data preparation step, the \gls*{stkg} construction.
\subsection{Spatio-Temporal Knowledge Graph Construction}
For the creation of our \gls*{stkg} we use Apache Sedona, which is an optimized data transformation engine for spatial data analysis. Overall, we base our \gls*{stkg} creation on the defined ontology we outlined in section \ref{cha:Ontology}. All aspects of the \gls*{stkg} construction are based on Apache Sedona functionality and use it as a backbone to provide a salable transformation engine. Our \gls*{stkg} creation is divided into four main parts: \textit{\gls*{osm} data tags}, \textit{\gls*{osm} geometry triples}, \textit{h3 grid cell triples}, \textit{h3 grid and \gls*{osm} geometry relation}.

As outlined in section \ref{cha:Ontology}, we separate our \gls*{osm} tags in two different categories. In the case where it is defined as a commonly used tag, we expose the key and value of the \gls*{osm} tags to individual entities. For all tags that do not fall into the category, we expose the key as a predicate and the value as the object. In addition, for all the respective we assign the date from the respective \gls*{osm} entity to our \gls*{stkg}, marking the respective date for the individual quadruple.

For the different parts, we follow the outlined approach from section \ref{cha:Ontology} where we expose for each geometry of an object in separate entities, storing them in the WKT format for our \gls*{stkg}. The storage of the geometry in the WKT format instead of the WKB format allows users that query the graph to be able to directly interpret the geometry of an object. Based on the respective elements for the grid, we model the neighborhood relationships for the grid cells with the same resolution. For grid cells that do not have the same resolution, the parent-child relationship is modeled and determined as outlined in section \ref{cha:Ontology}. For the relationships between the different \gls*{osm} geometries and the grid cell geometries, we use the predicate functions from Apache Sedona to determine the \gls*{de9im} predicates. With regard to the constructed \gls*{stkg} we outline in subsection \ref{cha:STKGCharacteristic} key statistics based on a selected set of osm.pbf files.

\subsection{Characteristics of constructed Spatio-Temporal Knowledge Graph} \label{cha:STKGCharacteristic}
For the creation of the \gls*{stkg} we provide a coding base for which we are able to scale spatial data analysis on large scale data representations. As a data base we use the yearly geofabrik data extracts from \gls*{osm}, which involves the datasets from the year 2018 to 2024. We use the .osm.pbf files from all continents which are provided on geofabrik. This involves the regions Africa, Antarctica, Asia, Australia, Central America, Europe, North America and South America \cite{boeckling_geofabrik_2024}. For the \gls*{osm} dataset in total this results in 529,065,633 distinct \gls*{osm} elements. For the h3 grid in total 3,675,984 individual grid cells are used in our \gls*{stkg}. In table \ref{tab:STKGStatistics} we outline the different key statistics for our \gls*{stkg}.

\begin{table}[ht]
    \centering
    \caption{Overview of Spatio-Temporal Knowledge Graph statistics}
    \label{tab:STKGStatistics}
    \begin{tabular}{p{0.45\textwidth}p{0.45\textwidth}}
    	\toprule
    	Metric name & Count\\
    	\midrule
    	Total triples & 27,042,753,856\\
        Distinct entity count & 1,841,912,579\\
        Distinct predicate count & 98,955 \\
    	\bottomrule
    \end{tabular}
\end{table}
\section{Conclusion and Outlook}
We have provided in our research a framework for the creation of \gls*{stkg} over the entire planet based on \gls*{osm}. Our data foundation builds up on a large representation of spatial data, representing various types of data. Representing those aspects in a \gls*{stkg} allows users to interact with changing spatial data over time. Similar to the outlined use cases of WorldKG or KnowWhereGraph, our \gls*{stkg} allows the possibility to explore connected entities based on shared metadata information or geographic relations.

In comparison to the presented \glspl{kg} in section \ref{cha:RelatedWork}, our approach provides a holistic representation of spatial data over the planet Earth. Compared to WorldKG, we provide with our \gls*{stkg} a representation of all available geometries in \gls*{osm}. Additionally, the usage of the \gls*{dgg} allows consumers to use the constructed \gls*{stkg} for various downstream tasks where the spatial grid is involved. Similar to KnowWhereGraph we allow consumers to make use of the individual \gls*{dgg} cell in their downstream tasks. Compared to both presented \glspl*{kg} our \gls*{stkg} has with over 27 billion triples and over 1,8 billion entities the largest coverage of spatial entities.

For our current implementation, we rely on third-party tooling provided by GDAL to transform the \gls*{osm} data format. While the module ogr2ogr shows a great efficiency in the transformation process, the handling of multiple large .osm.pbf files provides a certain overhead in the transformation phase. In a future implementation, the native support of .osm.pbf files in frameworks like Apache Sedona might help to reduce current initial processing times that are outside the data preparation related to the \gls*{stkg} construction. While \gls*{osm} provides a rich set of spatial data, it needs to be emphasized that \gls*{osm} data does not reflect the exact spatial reality and is also subject to vandalism.

With our approach, we have showcased that for the complete planet, all relevant geometries from \gls*{osm} and metadata. However, our current approach poses some limitations when it comes to traditional \gls*{kg} specific standards. We currently produce as an output file format delta files for our \gls*{stkg}. It provides for our large resulting \gls*{stkg} a good trade-off between the compression size of our data and the fulfillment of the ACID transaction consistency. However, \gls*{kg} specific frameworks like (Geo)SPARQL or are not directly supported in our file structure out of the box. There has been research around the mapping of SPARK SQL to SPARQL \cite{groth_sparqlgx_2016, schatzle_s2rdf_2015} or GeoSPARQL \cite{sattler_strabo_2022} to support efficient queries on large \glspl*{kg}. For future research, this could be evaluated compared to traditional \gls*{kg} frameworks that  \gls*{stkg}. In addition, in future research a comparison of the different \glspl*{stkg} on downstream spatial benchmark datasets could be performed to compare the different \gls*{stkg}.
\begin{acknowledgments}
  Map data copyrighted OpenStreetMap contributors and available from \url{https://www.openstreetmap.org}.
\end{acknowledgments}

\bibliography{bibliography}

\appendix

\section{Online Resources}

The documentation as also the coding for our research paper can be found on
\href{https://zenodo.org/doi/10.5281/zenodo.10857459}{GitHub}

\end{document}